Laurent Romary

# TEI and LMF crosswalks[1]

## Abstract

The present paper explores various arguments in favour of making the Text Encoding Initiative (TEI) guidelines an appropriate serialisation for ISO standard 24613:2008 (LMF, Lexical Mark-up Framework)[2]. It also identifies the issues that would have to be resolved in order to reach an appropriate implementation of these ideas, in particular in terms of informational coverage. We show how the customisation facilities offered by the TEI guidelines can provide an adequate background, not only to cover missing components within the current Dictionary chapter of the TEI guidelines, but also to allow specific lexical projects to deal with local constraints. We expect this proposal to be a basis for a future ISO project in the context of the on going revision of LMF.

Since this paper adopts the specific viewpoint of the TEI guidelines, no precise description of LMF is made here. For an introduction to LMF, see section 4 of (Romary 2013).

## 1 Towards a more intimate relationship between the TEI and the LMF standards

This chapter is about a simple thesis: the TEI framework could be the optimal serialisation[3] background for the LMF standard, since it provides both an ideal XML specification platform and a representation vocabulary that can be easily tuned (or *customized*) to cover the various LMF packages and components. This thesis does not come out of the blue but arises naturally when one observes the history of both initiatives, and their current impacts in various communities in the humanities and in computational linguistics, but also when one ponders on the relevance of having an LMF-specific serialisation when lexical data are in essence to be interconnected with various other types of linguistic resources.

As a matter of fact, the current XML serialisation of LMF suffers from both generic and specific problems that have prevented it from being widely used by the various communities interested in digital lexical resources. Right from the onset, the lack of consensus on the strategy to define a reliable and stable XML serialisation has forced the ISO working group on LMF to confine it to an informative annex, with the following main shortcomings:

Being carved in stone within the ISO standard, rather than being pointed to as an external and stable online resource, prevents it from being properly maintained, in order to either make corrections on identified weak points or bugs, or to add additional features;

It is only defined as a DTD, a vestigial XML schema language that hardly any XML developer currently uses anymore and which deeply limits its capacity to express constraints on types or to factorise global attributes. For the sake of simplicity (and this can be easily understood when one has to finalise a text for an ISO standard) no parallel definition of a RelaxNG or W3C schema was provided;

It does not reflect the intrinsic extensibility of LMF, as it does not contain any dedicated mechanism for customization, for instance when the developer of a new lexical model would like to discard some packages or add her own extensions;

A more intrinsic weakness of the suggested LMF serialisation is that it hardly takes up any existing vocabulary that could be reused to express either the macro- or micro-structure of a lexical entry. From a purely technical point of view, basic representation objects such as @xml:id or @xml:lang, which are standard practice in XML design, are redefined locally. At a low level, it misses using ISO 24610 for the representation of feature structures and redefines its own <feat> object[4]. As a whole, it suffers from a syndrome similar to that of the unfortunate ISO standard 1951[5]: it creates a specific silo that shows as little reuse of other initiatives as possible.

All in all, as we shall see in this paper, the TEI guidelines offer an appropriate answer to all the preceding issues. With a specification platform that allows the generation of multiple schema languages, a dynamic setting with short revision cycles, a proper integration of third party (ISO and W3C in particular) standards and of course the existence of a lexical representation basis with its *Dictionary* chapter, it provides the most flexible and reliable setting for deploying lexical applications that are meant to be compliant with the underlying LMF model.

Let us be clear: such infelicities as those we have notice above are usually the characteristics of standards that are in many other respects ahead of their time (think of ISO 8879:1986, SGML and its forerunner role for XML) and which require further years of ripening before they reach the best balance between comprehensiveness, simplicity and technical adequacy. The topic of our paper is indeed to contribute to improving LMF by considering bringing it closer to the TEI, an initiative that is well placed to demonstrate the importance of going through many years of fruitful iterations.

## 2    TEI as a data-modelling environment

Although the Text Encoding Initiative started nearly 3 decades ago in 1987, with its establishment as a consortium some 15 years ago, we will focus here on its current technical characteristics, knowing that the maintenance mechanisms we describe have contributed to its being the powerful infrastructure we know today.

The scope of the TEI mainly covers documents whose content can be seen as textual. This encompasses several possible object types such as manuscripts (BURGHART & REHBEIN 2012), scholarly papers (HOLMES & ROMARY 2010) or spoken data (SCHMIDT 2011). As we shall see lexical data are part of the covered domains but at this stage the most important feature to stress is that the almost 600 elements of the TEI guidelines are all defined in a specification language based on the TEI vocabulary itself. In a way, as was the case for Lisp[6] in the good old days, the TEI is expressed in its own language.

More fundamentally, the specification principles of the TEI infrastructure, reflected in the so-called ODD (One Document Does it all)[7] vocabulary, are based upon the concept of literate programming introduced by (KNUTH 1984), which advocates an integrated process through which technical specifications and prose descriptions are intimately linked with one another, so that one can easily work with one while having direct access to the equivalent object in the other. From the point of view of the TEI, this means that out of the ODD specification one can generate various schema formats (DTD, RelaxNG schemas, W3C schemas) as well as the documentation in any kind of possible format (pdf, docx, ePub, etc.).

Beyond the fact that the TEI is itself specified in ODD, the language is generic enough to be applicable to non-TEI environments. This has indeed been the case for several initiatives in the standardisation domain, the W3C using it for its ITS[8] recommendation, and ISO committee 37 using it for drafting several of its standards[9]. Moreover, ODD is well designed to combine heterogeneous vocabularies, like integrating CALS tables[10] or MathML[11] formulae within a TEI document. This is particularly important for the reuse of components (typically ISO-TEI feature structures) within a newly designed document model.

Without providing too many technical details here, we can describe the main aspects that give ODD its strength and flexibility:

The core declarative object is naturally the XML element, which can be associated with various descriptive properties (name, gloss, definition, examples and remarks) and technical information (content model based on RelaxNG snippets, further constraints (e.g. Schematron[12] rules), attribute declarations);

In complement to elements, the ODD language allows the definition of classes, which are grouping objects for elements having a similar semantics or occurring in the same syntactical context (for example all grammatical features). These are called *model classes*;

*Attribute classes* are also available to factorise attributes that are used uniformly by several elements (for instance all attributes providing additional temporal constraints to an element);

Elements may also be grouped together as *modules* (for instance: *drama*, *transcription of speech* and indeed *dictionaries*).

As described in (BURNARD & RAHTZ 2004) these various components provide a wealth of customization facilities, with for instance the possibility to add to or remove an element from a content model by changing its belonging to a given class in the TEI infrastructure. This specification and customization platform also paves the way to the description of coherent XML substructures (or *crystals*, ROMARY & WEGSTEIN 2012), that are essential for a component based data modelling and, as we shall see, correspond to the kind of granularity needed to implement LMF packages.

Finally, all these mechanisms are actually maintained and implemented as an open source portfolio of specifications[13] and tools[14] that facilitate their adoption by a wide range of users.

## 3    TEI as a quasi-LMF-compliant framework

Now that the motivations and general context for our approach have been set, we can focus on the actual representational tools that the TEI offers to deal with LMF compliant lexical structures. There are indeed two main approaches that one can consider here:

1. Considering lexical structures as **feature structures** and using the corresponding ISO-TEI joint vocabulary to this end;
2. Taking the XML vocabulary available from the **TEI chapter for dictionaries**.

### 3.1    The baseline – feature structures

The idea of representing lexical entries as feature structures has come to light in conjunction with the necessity of providing a structured representation of lexical data in the context of formal linguistic theories (POLLARD & SAG 1994; HADDAR et alii 2012 for an LMF proposal in this respect) but also to account for the deterministic representation and access to legacy dictionary data (VÉRONIS & IDE, 1992). As a matter of fact, since the early days of the TEI guidelines (LANGENDOEN & SIMONS 1995; LEE et alii 2004), there existed a specific module[15] inspired by these two trends and extensively covering all aspects of typed feature structures, with mechanisms for declaring constraints on them[16]. In 2006, following an agreement between the TEI consortium and ISO, the module became an ISO standard (ISO 24610-1) and is now the reference XML representation for feature structures.

Applying the ISO-TEI feature structure format for representing data in a way compliant to the LMF meta-model can be achieved quite straightforwardly by mapping LMF concepts as follows:

**Components** are implemented as features whose value is a complex feature structure;
**Elementary descriptors** (i.e. which correspond to *complex data categories* in the sense of ISO 12620) are implemented as elementary features with a symbolic value (mapped onto a *simple data category*).

Mappings between features and feature values with data categories can be controlled either by eliciting the association within a feature system declaration, or even by describing a feature library to factorise the information expressed within lexical entries. These mechanisms, related to the use of the so-called DCR[17] attributes (WINDHOUWER and WRIGHT 2012), are based upon the technical description provided in (ARISTAR-DRY et alii 2012) and will not be elaborated further here.

To visualize what such an LMF compliant representation could look like, we provide below a verbatim representation of the "clergyman" example from the LMF standard (cf. figure 4) according to the principles stated above[18].

```
<fs type="Lexicon" xmlns="http://www.tei-c.org/ns/1.0">

   <f name="language">en</f>

   <f name="LexicalEntry">

      <fs>

         <f name="partOfSpeech">commonNoun</f>

         <f name="Lemma">

            <fs>

               <f name="writtenForm">clergyman</f>

            </fs>

         </f>

         <f name="WordForm">

            <fs>

               <f name="writtenForm">clergyman</f>

               <f name="grammaticalNumber">singular</f>

            </fs>

         </f>
```

```
            <f name="WordForm">
                <fs>
                    <f name="writtenForm">clergymen</f>
                    <f name="grammaticalNumber"/>plural</f>
                </fs>
            </f>
        </fs>
    </f>
</fs>
```

**Example 1:** Inflected forms of clergyman represented as full feature structures

Even if one does not want to go as far as using fully-fledged feature structures but limits oneself to keeping at least the general principles of the LMF serialisation skeleton (elements named according to their equivalent component in the meta model), it is still possible to use the ISO TEI feature syntax for the corresponding descriptors in an LMF representation[19]. One possible advantage, beyond a better convergence across standardisation initiatives is that it allows, as was alluded to before, a simple declaration of the corresponding feature in connection to a data category registry such as ISOcat (WINDHOUWER & WRIGHT 2012). The suggested mixed-approach is illustrated below with the same "clergyman" example:

```
<LexicalResource xmlns:tei="http://www.tei-c.org/ns/1.0">
    <GlobalInformation>
        <tei:f name="languageCoding">ISO 639-3</tei:f>
    </GlobalInformation>
    <Lexicon>
        <tei:f name="language">eng</tei:f>
        <LexicalEntry>
            <tei:f name="partOfSpeech">commonNoun</tei:f>
            <Lemma>
                <tei:f name="writtenForm"/>clergyman</tei:f>
            </Lemma>
            <WordForm>
                <tei:f name="writtenForm">clergyman</tei:f>
                <tei:f name="grammaticalNumber">singular</tei:f>
            </WordForm>
            <WordForm>
                <tei:f name="writtenForm">clergymen</tei:f>
                <tei:f name="grammaticalNumber">plural</tei:f>
```

```
        </WordForm>
      </LexicalEntry>
    </Lexicon>
</LexicalResource>
```

**Example 2:** The clergyman example represented as a combination of LMF informative DTD and feature structures

All in all, the feature structure module of the TEI offers several possibilities to work within an LMF friendly environment, with the advantage of being based on a strong formalism where data validation is actually built-in. On the weak side, the generic character of feature structures, which comes with some degree of verbosity, makes it more difficult to maintain by human lexicographers but also provides less off-the-shelf validation facilities[20]. When this becomes an issue, it is reasonable to turn to a format that is natively intended to represent lexical structures such as provided by the dictionary module from the TEI.

### 3.2   The TEI *Dictionaries* chapter

The TEI guidelines actually come with a quite elaborate XML vocabulary for the description of electronic dictionaries[21]. Conceived initially on the basis of an underlying formal model of the hierarchical nature of a lexical entry (IDE & VÉRONIS 1995), and based upon previous theoretical (VÉRONIS & IDE 1992) and descriptive (IDE et alii 1992) works anticipating the idea of a solid structural skeleton further decorated by means of a variety of descriptors, it is not a surprise that the TEI model matches the LMF core package so well[22]. Still, it is important to keep in mind that the original chapter of the TEI guidelines, then named "Print dictionaries", was strongly oriented towards the representation of digitized material rather than on the creation of born digital lexical data. This had basically two consequences: a) it contains many more constructs intended for the representation of human oriented features (typically the etymology of a word (SALMON-ALT 2006; SALMON-ALT et alii-b 2005)) and b) it offers specific "flat" representations intended to cover the early steps of the digitization process, and that are outside the scope of the structured view we consider in this paper.

Whereas we will provide concrete crosswalks examples between the LMF model and the TEI *Dictionaries* chapter in the following section, we focus here on the description of the main elements that form the basis of the TEI descriptive toolbox for dictionaries.

The main structural elements of the TEI *Dictionaries* chapter are presented below and schematised in Figure 1 to illustrate their structural relationships:

**<entry>** is the basic structuring element of a lexicon (in the LMF sense) and groups together form information, grammatical information (cf. comments in the following section), sense information and related entries;

**<form>** can be used to describe one or several forms associated with an entry;

**<gramGrp>** groups together all grammatical features that may be attached to the entry as a whole (by means of its belonging to the *model.entryPart.top* model class) , to a specific form (through the *model.formPart* model class) or even as constraint on one of the senses of a word (again thourgh *model.entryPart.top*);

**<sense>** brings together all sense related information, i.e. definitions, examples, usage information and additional notes.

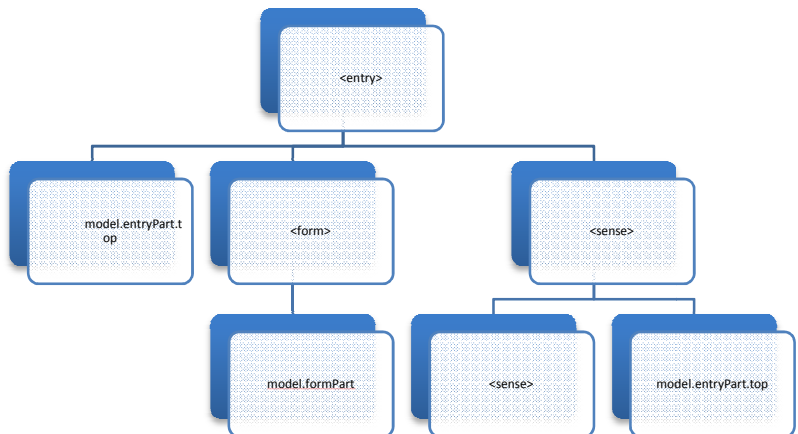

**Figure 1**: The simplified structure of an entry in the TEI *Dictionaries* chapter

The richness of the TEI descriptive toolbox has at times had the paradoxical effect that one could get deterred from using it simply because it does not come as a ready made module offering a single method of representing a given phenomenon. Although the same criticism could be addressed even more fiercely to the LMF standard itself, it is true that the experience gained over the years with the representation of lexical databases based on the TEI guidelines suggests that it is necessary to introduce more constraints, or at least some precise recommendation to make lexical representations more interoperable (cf. for instance ROMARY & WEGSTEIN 2012; BUDIN et alii 2012).

Among the core issues that sometimes make dictionary designers ponder upon which descriptive object to use is the variety of alternative elements that the TEI offers to <entry> proper. Apart from the possibility to group together homonyms (<hom>) or homographs (<superEntry>), the TEI has two specific elements for representing a lexical entry in a less structured manner: <entryFree> to allow any kind of combination and order of dictionary components, and <dictScrap>, which allows parts of a dictionary entry to be left un-encoded. These alternatives are indeed intended to deal with the specific scenarios of legacy human dictionaries, especially ancient ones, whose entries may not be straightforwardly organised (<entryFree>) or in the case of a multi-step scenario (<dictScrap>) whereby an initially OCRed dictionary is manually encoded step by step.

In the perspective of identifying the optimal customisation of the TEI guidelines that might implement the LMF model, we consider these various alternative constructs as transient objects that are part of specific workflows. For the purpose of disseminating LMF compliant data, we will thus from now onwards only consider <entry> as a proper implementation of the *LexicalEntry* component.

Another typical case of representational ambiguity results from the fact that the core sense-related sub-elements (<cit>, <def> or <usg>, with the ambivalent case of <gramGrp>) can actually occur freely as children of the <entry> element. This was initially intended to simplify representations where only one sense is being recorded and the encoder wants to avoid the supposedly superfluous <sense> element around such information. But at the end of the day, the resulting representations are not interoperable with one another and, in the context of the arguments made here, some of them are not even compliant with the LMF model. It is thus essential for the TEI community (or the LMF standard in one of its further revisions) to identify which subset of the TEI guidelines can be set as the reference LMF compliant one. As elicited in (ROMARY & WEGSTEIN 2012), such a customization should make <sense> mandatory for the representation of semantic content in <entry>, even if there is indeed only one sense.

Finally, on a more positive note, it can be observed that the TEI brings a lot of potential elements, which, in complement to the basic lexical encoding mechanisms provided by LMF, can be useful for the encoding of deep textual features with text fields. Typically, names, dates, foreign expressions in definitions or examples are not part of the LMF ontology. Still, they are usually important for the proper traversal or cross-linking of lexical material. Whether they are manually or automatically detected, the corresponding TEI vocabulary can definitely be used even as an external resource to LMF compliant representations[23] that are not expressed using the TEI guidelines proper. Typically a location can be tagged within a definition as in the following example:

```
<def>Orchidée épiphyte, originaire d'<geogName>Amérique tropi-
cale</geogName>, et dont l'espèce la plus connue est très recherchée
pour l'élégance de ses fleurs mauves à grand labelle en cornet on-
duleux.</def>
```

**Example 3:** Inline annotation of TEI content.

Such a wealth of inline annotation mechanisms should not be neglected when one is actually building up lexical resources from heterogeneous sources, which may actually contain such annotations (see for instance ECKLE-KOHLER et alii, 2012).

## 4 A canonical match: form representation in TEI

As we mentioned earlier, the *TEI Dictionaries* chapter already contains most of the basic constructs needed to implement the various components of the LMF core package. In this section, we would like to focus more specifically on the Form component and identify, a) how the available TEI elements for form description can be matched to the LMF specification and b) what perspective it brings about for the representation of full-form dictionaries, which we will take as an typical example of the type of lexical objects that are needed in the language technology domain (SAGOT 2010).

From an LMF point of view, the description of form information within a lexical entry (see figure 3) consists of a very simple, yet extremely expressive, structure based upon two components:

a **Form** component, which can be iterated within a lexical entry and unites all descriptions associated to what is considered as a single and coherent morphological object associated to the entry;

a **Form Representation** component, which allows one to provide as many descriptive views as needed for a given form.

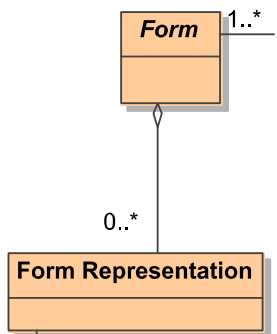

**Figure 2**: the Form and Form Representation components of the LMF core package

The two-level structure representation is an essential aspect to gain "form autonomy"[24] within a lexical entry. The canonical use of such a construct is typically when a word may occur in several written forms according to the script or transliteration mode being used. For instance, the Hangul representation of the verb "chida" (en: "to hit") can be associated with its Romanized transliteration as sketched below.

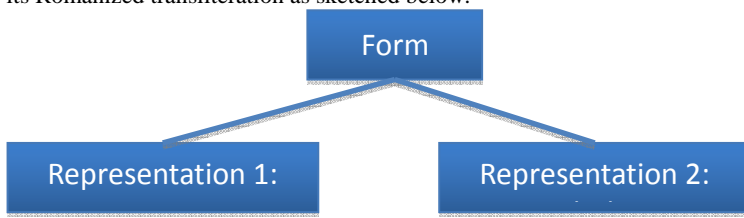

**Figure 3**: multiple scripting of the Korean verb "chida"

Given the canonical mapping that exists between the Form - Form Representation components in LMF and the <form> element - model.formPart model class in the TEI guidelines, this excerpt can be simply represented in TEI as follows, where the @xml:lang attribute is used to characterize the actual script (here, Hangul vs. Romanized) being used and the @type attribute provides some additional (e.g. project specific) categorisation of the corresponding linguistic segments.

```
<form>
    <orth type="standard" xml:lang="ko-Hang">치다</orth>
    <orth type="transliterated" xml:lang="ko-Latn">chida</orth>
</form>
```

**Example 4:** Multiple orthographic representations in TEI

If we now move to the slightly more elaborate "clergyman" example depicted in figure 4, the situation is hardly more complex and can be summarized by mean of the mapping table 1.

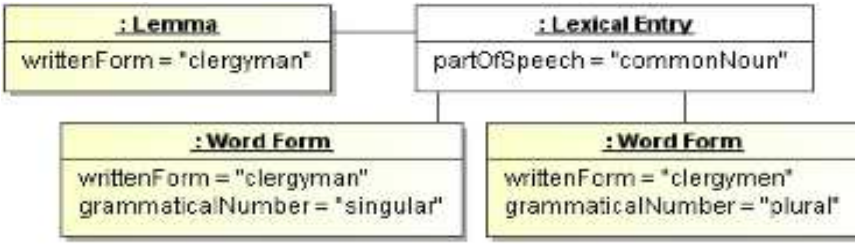

**Figure 4**: Schematic representation for the entry "Clergyman" (source: LMF standard)

| LMF component | TEI representation |
|---|---|
| LexicalEntry | \<entry> |
| Lemma | \<form type="lemma"> |
| Word Form | \<form type="inflected"> |
| writtenForm | \<orth> |
| partOfSpeech | \<pos> |
| grammaticalNumber | \<number> |

**Table 1**: Mapping between LMF components and corresponding TEI elements

The resulting representation, presented below, corresponds to a strict one-to-one mapping to the corresponding LMF model, which indeed can make it a strong basis for the implementation of any kind of full form lexica[25].

```
<entry>
    <form type="lemma">
        <orth>clergyman</orth>
        <gramGrp>
            <pos>commonNoun</pos>
        </gramGrp>
    </form>
    <form type="inflected">
        <orth>clergyman</orth>
        <gramGrp>
            <number>singular</number>
        </gramGrp>
    </form>
    <form type="inflected">
```

```
      <orth>clergymen</orth>
      <gramGrp>
         <number>plural</number>
      </gramGrp>
   </form>
</entry>
```

<p style="text-align:center;">**Example 5:** The clergyman example represented in compliance to the TEI guidelines</p>

As can be seen, the TEI guidelines provide quite a good coverage of the morpho-syntactic features typically needed for full form lexica. Still, there are several issues that have to be considered before one can systematically represent such lexica in an interoperable way for a variety of languages.

From a pure TEI point of view, we already tackled the issue of representational ambiguity, which can make encoders use different constructs to represent the same phenomenon. In the case of inflected forms, both the coherence of their representation and the necessity to remain compliant with LMF requires a systematic use of <form> and <gramGrp> to embed form and grammatical related information respectively, even if in both cases it may be seen as redundant. In the preceding example for instance, even if only a single grammatical feature (<number>) appears in the <gramGrp>, a coherent representation with other word categories (for instance verbs) or other languages, requires that the latter should not be omitted[26]. This allows for instance that a search for the various grammatical constraints used in a lexicon can be made with <gramGrp> as an entry point.

From a data model perspective, this also ensures, as demonstrated in the previous section, a coherent and strict equivalence of <gramGrp> with a feature structure in case one wants to use this generic representation means in place of <gramGrp> within <form>. For instance, the previous example can be reformulated as[27]:

```
<entry>
   <form type="lemma">
     <orth>clergyman</orth>
     <fs type="grammar">
       <f name="pos">commonNoun</f>
     </fs>
   </form>
   <form type="inflected">
     <orth>clergyman</orth>
     <fs type="grammar">
       <f name="number">singular</f>
     </fs>
   </form>
```

```
  <form type="inflected">
    <orth>clergymen</orth>
    <fs type="grammar">
      <f name="number">plural</f>
  </fs>
  </form>
</entry>
```

**Example 6:** The clergyman example represented in compliance to the TEI guidelines with feature structures

Finally, we should address here the issue of linguistic coverage, with the possibility of constraining the semantics of the grammatical features used in such representations, and furthermore to add features that may not be part of the core grammatical elements of the TEI, but which are still necessary to describe morpho-syntactic constraints in other languages. For this purpose, the TEI provides a generic <gram> element, which, coupled with the appropriate value for its @type attribute, can theoretically mark any kind of grammatical feature. Still, it is strongly recommended, when one has such a representational need, to design an *ad hoc* element in one's ODD specification and relate this specification to ISOcat by means of either the <equiv> construct or the appropriate DCR attributes[28].

## 5    Adding components to the TEI framework: the syntactic case

Since the *TEI Dictionaries* chapter was initially conceived to account for the kind of information that appears in machine-readable dictionaries, it only sparsely covers features related to language processing and in particular does not propose any specific element for representing syntactic or semantic structures. When one looks at the various additional packages of LMF on the one hand and at the customisation facilities of the TEI infrastructure on the other, it appears to be relatively easy to define extensions that actually allow TEI based customisation to include the missing LMF constructs.

In this section we present the basic principles to be applied to create such a customization that extends the TEI guidelines by means of an ODD specification for the syntactic package of LMF. This presentation will be carried out by going through a specific example, namely the encoding of verbal structures in CoreNet, the Korean Wordnet.

CoreNet, the Korean Wordnet lexicon (also known as CoreNet, see (CHOI 2003) and (CHOI et alii 2004)) has been put together as a deep semantic and syntactic encoding of a selection of the 50 000 Korean most frequent words (mainly nouns and verbs). Looking at verbs proper, their representation is based upon a double filing system of a) *verb concepts*, associating a concept number (and therefore a Wordnet Synset, via a specific conceptual mapping) to the various senses and b) *verb frames*, associating each sense with one or several predicate-argument structure.

**Figure 5**: An entry from the verb concept section of CoreNet (senses are marked in red, sub-senses in green)

As illustrated in figure 5 for the verb "chida" (치다), the verb concept structure is organised in senses and sub-senses, to which are attached both a Wordnet reference and a gloss. This two-level semasiological representation is indeed entirely construable as a standard TEI <entry> structure as illustrated below:

```
<entry>
   <form>
      <orth type="한글">치다</orth>
      <orth type="Romanization">chida</orth>
   </form>
   …
   <sense n="3">
      <gramGrp>
         <subc>vt</subc>
      </gramGrp>
```

```
      <sense n="1">
         <ref type="wordnet">
            <idno>1221282691</idno>
            <gloss>치기</gloss>
         </ref>
      </sense>
      <sense n="2">…
      </sense>
   </sense>
</entry>
```

**Example 7:** Partial TEI representation of an entry from CoreNet (*chida*)

The verb-frame structure is in turn illustrated in figure 6, where one can see that a complementary semasiological structure is being used, grouping together senses from the verb concept structure (represented here by a combination of concept number and gloss) and associating such groups to one or several predicate-argument representations. An additional Japanese gloss is provided for each semantic group, on the basis of the actual semantic restriction introduced for the corresponding arguments.

**Figure 6**: Two entries from the verb frame section of CoreNet

This predicate argument structure is indeed a good instance of the syntactic extension of LMF, which is based on the notion of a sub-categorisation frame (component: Sub-categorisation Frame), which in turn is linked to various syntactic arguments (component: Syntactic Argument). Figure 7, which takes up an Italian example from the LMF standard, illustrates this core structure and shows how it is directly anchored on the Lexical Entry component.

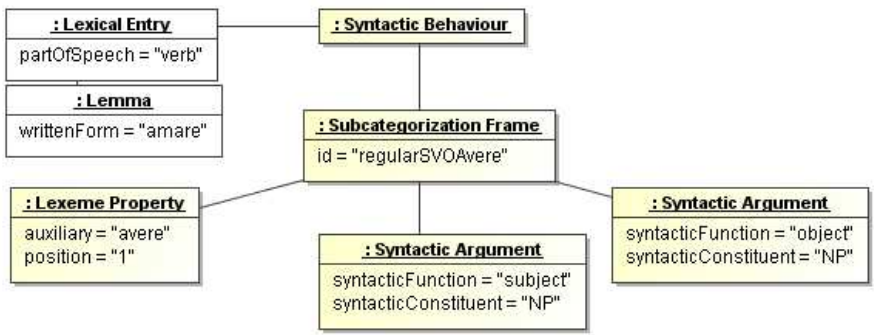

**Figure 7**: An instance of the LMF syntactic extension (source ISO 24613)

When transposing this model to our CoreNet example, we can actually embed the syntactic description within the sense level of the lexical entry[29]. This leads to a possible TEI extended construct that may look as follows:

```
<tei:sense>

    <tei:gloss xml:lang="ja">ふぶく</tei:gloss>

    <lmf:syntacticBehaviour>

        <lmf:subcategorizationFrame>

            <lmf:syntacticArgument>

                <lmf:syntacticFunction>N1</lmf:syntacticFunction>

                <tei:colloc type="particle" xml:lang="ko">

                    이/가</tei:colloc>

                <tei:gloss xml:lang="ko">눈보라</tei:gloss>

                <tei:ref type="wordnet">

                    <tei:idno>12231214</tei:idno>

                    <tei:gloss xml:lang="ko">눈</tei:gloss>

                </tei:ref>

            </lmf:syntacticArgument>

        </lmf:subcategorizationFrame>

    <lmf:syntacticBehaviour>

</tei:sense>
```

**Example 8:** Inclusion of a syntactic construct in the TEI representation of an entry from CoreNet (*chida*).

In this representation, we applied the following core specification principles, which, to our view, should be systematically applied for any further TEI based LMF extension:

Limit the introduction of specific elements to those for which there are no equivalent constructs in the TEI infrastructure

Keep new elements within their own namespace. This is a general principle for TEI customization, but it allows here a clear management of the heterogeneous mix-up of elements that we suggest here at all levels of the representation

Avoid introducing new LMF elements within existing TEI constructs apart from the clear anchoring of the LMF syntax crystal within the <sense> element. This principle is essential to facilitate the future integration of our proposal as an official extension to the TEI guidelines, where unintended side effects should be avoided

As a side note, we can see the interesting case of the various usages of the TEI <gloss> element in this representation. Depending on the context, it can be applied in a systematic way to mark any kind of equivalent wording in the various object or working languages of the dictionary.

The actual implementation of such an extension is rather straightforward. Following the general principles outlined in (TBE 2010) for implementing a TEI customisation in ODD, we only give here the essential aspects of the proposed syntax extension to the TEI Dictionaries chapter[30].

The first step is to create a background customisation comprising the core modules of the TEI guidelines together with the Dictionaries module as follows:

```
<schemaSpec ident="LMFSyntax">
    <moduleRef key="core"/>
    <moduleRef key="tei"/>
    <moduleRef key="header"/>
    <moduleRef key="textstructure"/>
    <moduleRef key="dictionaries"/>
</schemaSpec>
```

**Example 9:** Outline of the ODD specification TEI customisation for dictionaries.

The second step is to create specifications for all new elements within a specific LMF namespace. When such elements have a complex content model, an associated element class is created so that the content model is easy to customise further. For instance, a simplified specification for the <syntacticArgument> element may look as follows:

```
<elementSpec ident="syntacticArgument" module="Syntax"
    ns="http://www.iso.org/ns/LMF">
    <classes>
        <memberOf key="model.subcategorizationFramePart"/>
```

```
   </classes>
   <content>
      <rng:oneOrMore>
         <rng:ref name="model.syntacticArgumentPart"/>
      </rng:oneOrMore>
   </content>
</elementSpec>
```

**Example 10:** ODD specification for the <syntacticArgument> element.

Finally, as seen also in the preceding example each element is made a member of the appropriate classes to appear in the intended content models.

The resulting specification is all in all quite simple and allows one to edit syntactic lexica right away, while remaining within the TEI realm. Moreover, it shows that implementing similar extensions for some additional packages would definitely be an easy tasks that would not take too much time for a minimally TEI minded person.

## 6 Contributing to the LMF packages: linguistic quotations

We now address the opposite case to the one we have just seen, namely when some existing constructs in the TEI infrastructure do not have any counterpart in the LMF standard and can thus contribute to defining additional packages. There are indeed several such interesting cases in the TEI guidelines (one may think in particular of all etymological related aspects), but in order to make the point clear we will focus on a simple yet essential type of information: *quotation structures*.

Quotations in a lexical database are linguistic segments that illustrate the use of the headword either as a constructed example, as the citation of an external source or through the embedding of excerpts that have been automatically extracted and selected from a corpus. In some lexicographic projects (cf. e.g. KILGARRIFF & TUGWELL 2001 or SINCLAIR 1987) such quotations have even been the organising principle of the whole lexical matter.

In their simplest form, quotations appear as a textual sequence embedded within other descriptive information of the word, for instance[31]:

> **ain't (eInt)** *Not standard. contraction of* am not, is not, are not, have not or has not: *I ain't seen it.*

When the quotation is actually taken from a known source, it is usually accompanied by an explicit (usually abbreviated) reference to it, as in[32]:

> **valeur** … n. f. … 2. Vx. Vaillance, bravoure (spécial., au combat). 'La valeur n'attend pas le nombre des années' (Corneille).

In the case of multilingual dictionaries, we can extend the notion of quotations to the provision of a translation, possibly accompanied by additional contextualising information. This falls indeed within our earlier definition of a quotation, since such translations actually illustrate the intended meaning in the target language. In the following example we see for instance how such a translation can in turn be refined by an explicit gloss for the corresponding meaning:

> **rémoulade [Remulad]** nf remoulade, rémoulade (*dressing containing mustard and herbs*).

Further types of quotation refinements can be observed in existing dictionaries and indeed, any kind of morpho-syntactic, syntactic or semantic information may be associated with quotations, as long as it provides a qualification for the corresponding usage. Taking again the case of multilingual dictionaries, it is indeed standard practice to refine a translation by means of gender information as in the following excerpt:

> **dresser** … (a) (Theat) habilleur m, -euse f; (Comm: window ~) étalagiste mf. she's a stylish ~ elle s'habille avec chic; V hair. (b) (tool) (for wood) raboteuse f; (for stone) rabotin m.

In this example, we see various types of refinements, with a simple marking of gender for the translation (*habilleur m*), to a combination of morpho-syntactic and semantic constraints (*(for wood) raboteuse f*).

As can be seen, quotation structures are a strong component of the organisation of lexical entries in senses. We are used to observing these in traditional print dictionaries, but indeed, it is easy to foresee a generic mechanism that applies to any lexical database where illustrative text (examples or translations) are to be integrated.

In this respect, the TEI has taken this issue very seriously by introducing in its recent editions (from P5 onwards), a single construct based on the <cit> element[33] that merged the various specific constructs that existed for examples (the <eg> element in the P4 edition of the TEI guidelines) or translations (the <tr> element in P4). This construct can be characterised as follows:

it is based upon a very generic two-level structure where the <cit> element is the entry point and comprises a language excerpt expressed by means of a <quote> (occasionally a <q>) element;

the <cit> element may have a @type attribute to further constrain the nature of the quotation construct, for instance "example" or "translation".

In the simplest case, when no further constraint or bibliographic reference is needed, the <cit> construct boils down to something as simple as the following example representing a translation[34]:

```
<cit type="translation" xml:lang="fr">

   <quote>horrifier</quote>

</cit>
```

<div align="right">

**Example 11:** Simple example for <cit>.

</div>

When further refinements are expressed in relation to the quotation, these are added to the actual quoted sequence, using the usual descriptive vocabulary available from the TEI guidelines. For instance, the provision of the gender for the French equivalent to the headword "dresser" in English would be expressed as follows:

```
<cit type="translation" xml:lang="fr">

   <quote>habilleur</quote>

   <gramGrp>

      <gen>m</gen>
```

```
    </gramGrp>
</cit>
```

**Example 12:** <cit> construct with grammatical constraints.

Finally, an important feature of the <cit> element is its recursivity where for instance the actual translation for a example is also provided, as in the following example:

```
<cit type="example">
    <quote>she was horrified at the expense.</quote>
    <cit type="translation" xml:lang="fr">
        <quote>elle était horrifiée par la dépense.</quote>
    </cit>
</cit>
```

**Example 13:** <cit> construct with a translation of the main example.

The LMF standard does not have a real equivalent to the <cit> crystal and the only similar structure that appears in LMF may be the possibility to associate a statement in a definition[35]. We thus propose to define an optional extension to the LMF core package, anchored on the sense component and schematized in figure 8.

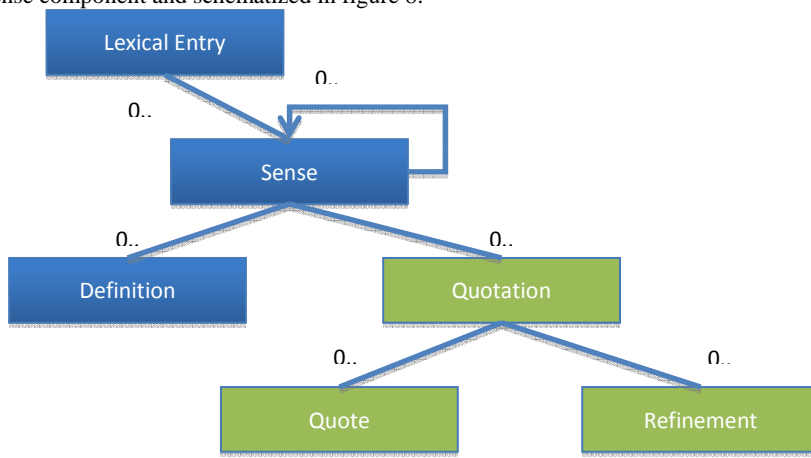

**Figure 8**: A sketch for a possible Quotation package in LMF

As we can see, the package is directly part of the Sense component aggregation and further defined as a combination of a Quote (an instance of the Text Representation component in LMF) and a Refinement component.

A further specification process, which should be carried out in consultation with the community of lexical databases developers and users, should clarify what should pertain to the Refinement component in this model. As we have seen, we have here a wide spectrum of possibilities, ranging from authorship or bibliographical information to morpho-syntactic constraints and comprising various alternative forms (pronunciation, variants, translations) or usage information (subject, definition, gloss). Of course, a possible instance of a Requirement may also be a Quotation.

## 7 Towards more convergence between initiatives: a roadmap

One of the underlying aims of this paper is to demonstrate that there are some good possibilities to work towards a better convergence between the LMF and the TEI initiatives in the domain of lexical structures, and in particular take full benefit of each side's strengths. Indeed, whereas the ISO perspective brings stability and an international validation, it should not be neglected how large the current TEI community is. With this perspective in mind, the project of having an LMF serialisation entirely expressed as a TEI customisation can be seen as a most important endeavour to offer a common and strong basis for any kind of lexical work both in the language technology and the digital humanities domains. This will also provide LMF with a real customisation platform that will facilitate the work of defining project specific subset within a coherent framework that guaranties compliance to the underlying reference standard.

There is indeed a good window of opportunity to go in this direction. ISO committee TC 37/SC 4 has issued a plan in 2015 to revise ISO standard 24613 so that it becomes a multipart standard reflecting the variety of domains addressed so far within one single document. In this context, it would probably be appropriate to submit a specific part dedicated to the serialisation of LMF by means of the TEI guidelines on the basis of the principles expressed in this paper. Even if we cannot anticipate, at the time of publication of this paper, the possible success of such an endeavour, the various positive signs received already by the author of this paper are encouraging to carry this out as far as possible.

## Acknowledgements

I want to address here my deep and friendly thanks to the colleagues and friends who provided such a valuable feedback on early draft of the paper, in particular JUDITH ECKLE-KOHLER, MARTIN HOLMES, JOHN MCCRAE, CHARLY MÖRTH, DANIEL STOEKL, TOMA TASOVAC, WERNER WEGSTEIN, MENZO WINDHOUWER as well as the two anonymous reviewers of JLCL.

## Abbreviations

| | |
|---|---|
| DCR | Data Category Registry |
| FSD | Feature System Declarations (ISO 24610-2:2011) |
| ISO | International Organisation for Standards |
| LMF | Lexical Markup Framework (ISO standard 24613:2008) |
| ODD | One Document Does it all, the specification subset of the TEI guidelines |
| RelaxNG | Regular Language for XML Next Generation |
| SGML | Standard Generalized Markup Language  (ISO 8879:1986) |
| TEI | Text Encoding Initiative |
| W3C | World Wide Web Consortium |
| XML | Extensible Markup Language (W3C recommendation) |

¹ Most abbreviations are elicited when they appear for the first time in the text. A complete abbreviation section is available at the end of this paper, right before the bibliographical references.

² We will henceforth refer to the ISO document as simply *LMF*.

³ "Serialisation" means a concrete data representation on computers for the sake of storage or interchange. A serialisation, for instance an XML format, is often conceived in compliance with a reference model (in the case of our paper, LMF).

⁴ The LMF <feat> object is not even compliant with ISO standard 16642 (TMF) which defined such an element before ISO 24610 was in place.

⁵ See (LEMNITZER et alii 2013) for a more precise analysis of the difficulties related to ISO 1951.

⁶ See *https://en.wikipedia.org/wiki/Lisp_(programming_language)*

⁷ See a technical introduction in *http://www.tei-c.org/Guidelines/Customization/odds.xml*

⁸ Internationalization Tag Set; *http://www.w3.org/TR/its/*

⁹ ISO 24611, ISO 24616, ISO 24617-1, and on going revision of ISO 16642

¹⁰ Maintained by OASIS, see *https://www.oasis-open.org/specs/tablemodels.php*

¹¹ Maintained by W3C, see *http://www.w3.org/Math/*

¹² *http://www.schematron.com*

¹³ *https://github.com/TEIC/TEI*

¹⁴ For instance, Roma (*http://www.tei-c.org/Roma/startroma.php*) for the online design of customization, or Oxgarage (*http://www.tei-c.org/oxgarage/*) for the transformation of TEI documents from and to various possible formats or schema languages.

¹⁵ Chapter 18 in TEI P5 - *http://www.tei-c.org/release/doc/tei-p5-doc/en/html/FS.html*

¹⁶ FSD – Feature System Declarations

¹⁷ Data Category Registry

¹⁸ In all our examples, we will use the simplified (untyped) form for feature values as plain text content of the <f> element. More elaborate implementations should distinguish specific subtypes as specified in the ISO-TEI specification.

¹⁹ A very similar approach has indeed been developed by MENZO WINDHOUWER in the context of the RELISH project, see *http://tla.mpi.nl/relish/lmf/* and (ARISTAR-DRY et alii 2012)

²⁰ Note that the same criticism applies to RDF based representations, which should only be contemplated for some specific end-user delivery scenarios.

²¹ see *http://www.tei-c.org/release/doc/tei-p5-doc/en/html/DI.html*

²² It is even less surprising given that the TEI principles informed the first ISO meeting in Korea (February 2004) where the first LMF consensus was put together (ROMARY et alii 2004)

²³ see for instance the chapter "Names, Dates, People, and Places" (*http://www.tei-c.org/release/doc/tei-p5-doc/en/html/ND.html*) for the encoding of basic name entities.

²⁴ Like we have the term autonomy principle in terminology

[25] See also the first experiments done on the Morphalou dictionary (ROMARY et alii, 2004) or for the Arabic language (SALMON-ALT et alii, 2005-a)

[26] In the case where there is no grammatical information available, the <gramGrp> element should be of course omitted. Indeed, it is important to keep to the general encoding rule of avoiding the insertion of useless void elements (With thanks to MARTIN HOLMES for pointing this out to me).

[27] CHARLY MÖRTH rightly mentions that when implementing such a solution on a large scale it may be appropriate to move all <fs> elements into <fLib> elements and use an @ana attribute on form to refer to them.

[28] Namely: dcr:datcat and dcr:valueDatcat

[29] The full LMF package for syntax is (rightly) intended to allow the factorisation of syntactic constructs across several entries. We simplify the representation here to make our point clearer. The full ODD specification should indeed implement both views.

[30] The complete customisation is available under *http://hal.inria.fr/hal-00762664*

[31] Source: TEI P5, chapter "Dictionaries", *http://www.tei-c.org/release/doc/tei-p5-doc/en/html/DI.html* (original source: *Collins English Dictionary*. London: Collins)

[32] *ibid.* (original source: GUERARD, FRANÇOISE (1990). *Le Dictionnaire de Notre Temps*, Hachette, Paris)

[33] *http://www.tei-c.org/release/doc/tei-p5-doc/en/html/ref-cit.html*

[34] We recommend this construct rather than the simpler: <gloss xml:lang="en">horrifier</gloss>, with the assumption that it is better to use the same structure (<cit>) for both glosses and illustrative quotations. Thanks to Martin Holmes for pointing to this.

[35] cf. ISO 24613 "*Statement* is a class representing a narrative description and refines or complements *Definition*."